\documentclass[letterpaper, 10 pt, conference]{ieeeconf}  

\IEEEoverridecommandlockouts                              

\usepackage{cite}
\usepackage{amsmath}
\usepackage{amsfonts}
\usepackage{amssymb}
\usepackage{bbm}
\usepackage{graphicx}
\usepackage{url}
\usepackage{subcaption}
\usepackage[export]{adjustbox} 
\usepackage{booktabs}
\usepackage{multirow} 
\usepackage{xcolor}

\definecolor{mmcolored}{rgb}{1.0,0.0,0.0}

\title{\LARGE \bf
MOSAIC: \underline{M}odular \underline{O}bstacle and \underline{S}patial-\underline{A}ware Framework for Manipulation \underline{I}n \underline{C}luttered Scenes
}

\author{Richard Shi$^{1,2*}$, Mohammad Mahdavian$^{1*\dagger}$, Mehrdad Momeni Zadeh$^{1,3*}$, Mohammed Elmahgiubi$^{1}$, Amir Rasouli$^{1}$ 
\thanks{*Equal Collaborations}
\thanks{$\dagger$ Project lead and corresponding author\newline
{\tt\small\{mohammad.mahdavian1@huawei.com\}}
}
\thanks{$^{1}$Huawei Noah's Ark Lab , $^{2}$University of Toronto , $^{3}$Simon Fraser University
    }%
}

\begin{document}

\bstctlcite{IEEEcontrol}

\maketitle
\thispagestyle{empty}
\pagestyle{empty}

\begin{abstract}

Vision-Language-Action (VLA) models have demonstrated strong performance in robotic manipulation, yet most rely primarily on 2D visual observations and language instructions, leaving scene geometry and obstacle relationships to be inferred implicitly. This limitation is particularly problematic in cluttered environments, where small spatial errors can lead to unintended contacts or collisions. We propose MOSAIC, a modular spatial- and obstacle-aware framework that can be integrated with pretrained VLA policies to improve object-level scene understanding and action generation. Our method augments the VLA backbone with an auxiliary semantic grounding task that predicts role-specific masks for the target object, destination region, and surrounding obstacles. We further combine the target and destination masks with depth estimated from 3D scene reconstruction to construct role-specific geometric tokens that are injected directly into the action expert. In addition, we propose obstacle-aware attention supervision that emphasizes obstacles near the robot body during task execution. The resulting attention features are subsequently injected into later action-expert layers to further guide action generation. Together, these components provide explicit semantic, geometric, and obstacle-aware cues while preserving the underlying pretrained VLA architecture. We evaluate our framework on pretrained VLA model using the RoboPRO benchmark, including visual perturbation tests, and further validate it on real-world manipulation tasks. MOSAIC improves the success rate by 5.1\% in cluttered scenes, and by 9.7\% under visual perturbations, with gains of up to 25.0\% in the most challenging environment. We will release our code and dataset upon acceptance at the following Github page: \url{https://mmahdavian.github.io/MOSAIC/}

\end{abstract}

\section{INTRODUCTION}

VLAs have emerged as powerful robotic foundation models for understanding task context, spatial reasoning, and executing physical-world tasks. VLAs typically build on Vision-Language Models (VLMs)~\cite{beyer2024paligemma,yang2025qwen3} by extending their interfaces with proprioceptive inputs and action outputs. Actions are commonly generated either through discrete token prediction~\cite{brohan2022rt,brohan2023rt} or specialized action experts based on diffusion~\cite{ho2020denoising,team2024octo} or flow matching~\cite{lipman2022flow,black2024pi_0,intelligence2025pi_}. Despite their strong performance, generalization to unseen scenarios remains challenging, particularly for manipulation in cluttered scenes~\cite{rasouli2026distracted}.

Several approaches improve VLA generalization through large-scale pretraining~\cite{kim2024openvla}, cross-embodiment learning~\cite{o2024open}, and world or future-state modeling~\cite{wu2024unleashing}, however at the cost of substantial data and computation. Recent methods incorporate explicit spatial grounding through 3D geometry~\cite{qu2025spatialvla} or task-relevant objects understanding~\cite{huang2025roboground,vo2025clutter}, providing richer scene representations for more generalizable manipulation. However, these approaches generally do not explicitly model surrounding obstacles as a separate component.

To improve object understanding of VLA models, existing approaches condition the policy on object-level 2D information, such as target-object masks~\cite{tu2026sg} or target and destination regions~\cite{huang2025roboground}, or inject depth and 3D point-cloud information directly into the action expert~\cite{sun2025geovla,li2026pointvla,vo2025clutter}. While such grounding improves the spatial representations available for action generation, these methods at best focus on the target object and its destination. In cluttered environments, the lack of explicit obstacle awareness can lead to unintended contacts, collisions, and ambiguous actions. 

A further challenge caused by object clutter and visual noise is that they can divert the policy’s attention away from task-relevant regions. Prior work shows that VLA action experts naturally attend to the target object and destination~\cite{park2026your}. Several methods further guide this attention by aligning model attention maps with human gaze estimates or target-object masks to improve performance and interpretability~\cite{pani2026gaze,jia2026guidedvla}. But these approaches rarely guide attention toward surrounding obstacles, which are critical for safe and reliable manipulation in cluttered scenes.

\begin{figure}[t]
    \centering
    \includegraphics[width=\linewidth, trim=0 0cm 0cm 0cm, clip]{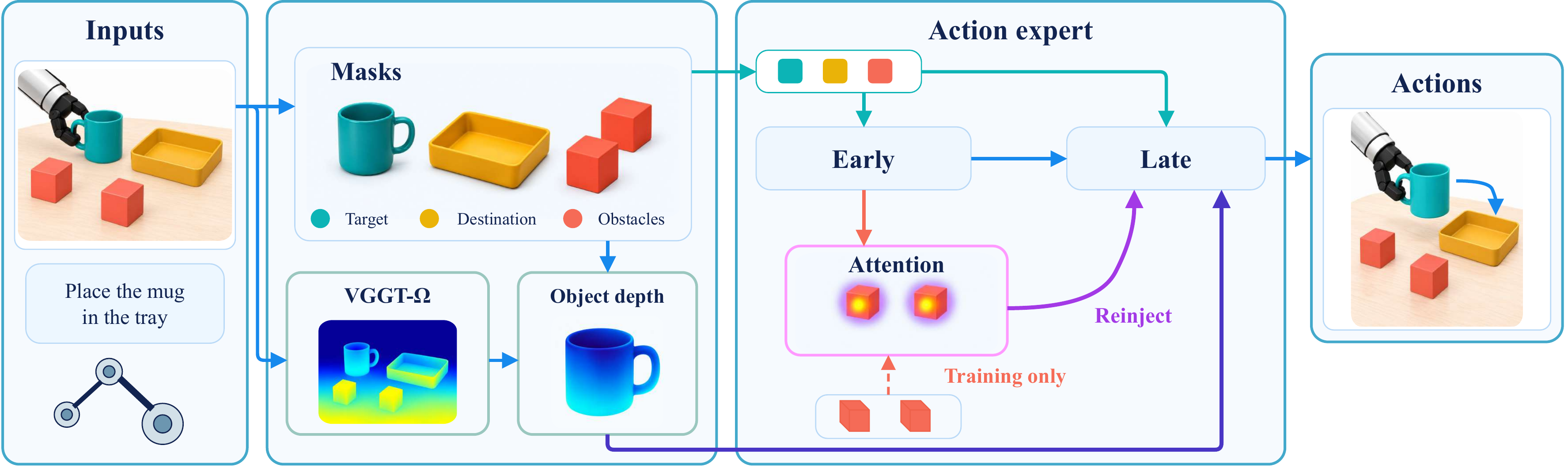}
    \caption{\textbf{MOSAIC overview}. A pretrained VLA is augmented with role-specific masks, object-level depth tokens, and obstacle-focused attention guidance.}
    \label{fig:general} \vspace{-5mm}
\end{figure}


To improve spatial and object-level reasoning in VLA models, we propose MOSAIC (Fig.~\ref{fig:general}), a modular framework that combines semantic masks from the VLA backbone with VGGT-$\Omega$~\cite{wang2026vggt} to extract role-specific object geometry. These 3D features are encoded as tokens and injected into the action expert for clutter-aware manipulation. We further introduce contact-point-aware attention supervision over the obstacles, and re-inject the resulting attention features into later action-expert layers to guide action generation. 

In summary, our main contributions are as follows:

\begin{itemize}
\item We propose MOSAIC, a novel modular framework that enhances object-centric understanding and spatial reasoning in VLA models, enabling more effective manipulation in cluttered environments.

\item Our modules augment a pretrained VLA with components that explicitly model surrounding obstacles using its own backbone, while injecting semantic object-level depth information into the action-generation process.

\item We guide VLA attention beyond the target object and destination region toward surrounding obstacles, improving action generation in cluttered scenes.

\item We demonstrate improvements in manipulation performance when applying our proposed modules to pretrained VLA models, evaluated on the RoboPRO benchmark and experimented with real-world robots.

\end{itemize}

\section{Related Work}

\subsection{Vision Language Action Models}
VLA models generate robotic actions from visual observations and natural-language instructions, typically using pretrained VLMs as their backbone. Early approaches formulate action prediction as next-token generation~\cite{brohan2022rt,brohan2023rt}, while more recent methods employ dedicated action experts trained with diffusion~\cite{ho2020denoising} or flow-matching~\cite{lipman2022flow} objectives~\cite{black2024pi_0,intelligence2025pi_,intelligence2025pi,intelligence2026pi,bjorck2025gr00t}. By decoupling action generation from language tokenization, these models achieve stronger downstream performance. However, most still rely primarily on 2D visual inputs and language, leaving depth, object proximity, and free space to be inferred implicitly. This is particularly limiting in cluttered scenes, where inaccurate spatial reasoning can lead to unintended contacts or collisions with nearby obstacles.

\subsection{Visual Grounding for VLAs}
Recent approaches incorporate spatial reasoning for VLAs inspired by methods developed for VLMs~\cite{chen2024spatialvlm,cheng2024spatialrgpt,schofield2026chain}. RoboGround~\cite{huang2025roboground} extracts 2D masks for the target object and destination region and conditions autoregressive action generation on this grounded visual information. OBEYED-VLA~\cite{vo2025clutter} similarly uses multi-view masks together with depth cues to identify task-relevant objects and improve action generation. However, RoboGround, OBEYED-VLA, and related methods~\cite{tu2026sg,li2025controlvla,zuo2026gaze2act} mainly condition on the target object and, in some cases, the destination region, without explicitly modeling surrounding obstacles. 

A complementary direction is to incorporate 3D geometry directly into action generation. GeoVLA~\cite{sun2025geovla} and PointVLA~\cite{li2026pointvla} inject full-scene point clouds, while OCRA~\cite{wang2026ocra} and SAM3D-VLA~\cite{liu2026sam3d} use object-level geometric features. However, OCRA does not explicitly preserve task-specific object roles, while SAM3D-VLA distills SAM3D~\cite{chen2026sam} features into the VLA backbone and learns the semantics of the target object only implicitly. In contrast, our method explicitly separates the target, destination, and obstacle roles, represents each with dedicated geometry tokens, and injects the target- and destination-specific tokens directly into the action expert.

\subsection{Attention Guidance for VLAs}
Attention guidance can improve both model performance and interpretability by directing the model toward task-relevant regions during training or inference. KNOWS~\cite{park2026your} proposes a training-free safety filter using attention heads that localize the target object, while other methods often introduce training-time supervision to align attention with target regions~\cite{pani2026gaze,jia2026guidedvla,xiao2026ava,song2026reconvla,li2026gazevla}. These approaches typically rely on ground-truth attention maps or gaze-related auxiliary objectives. For example, Gaze-Regularized VLA~\cite{pani2026gaze} supervises attention-map alignment and motivates our heatmap construction. However, most existing methods focus primarily on target regions and the robot end effector, without explicitly considering nearby obstacles or the full robot body. In our approach, we interpolate points along the robot links to capture full-body obstacle proximity and potential contact locations. These signals are used to construct fine-grained ground-truth attention distributions for supervising the action expert. The resulting attention maps are then injected into the action tokens through a residual connection, enabling more obstacle-aware action generation in cluttered environments.

\begin{figure*}[t]\vspace{2mm}
    \centering
    \includegraphics[width=\linewidth,  trim=0 0cm 0 0cm, clip]{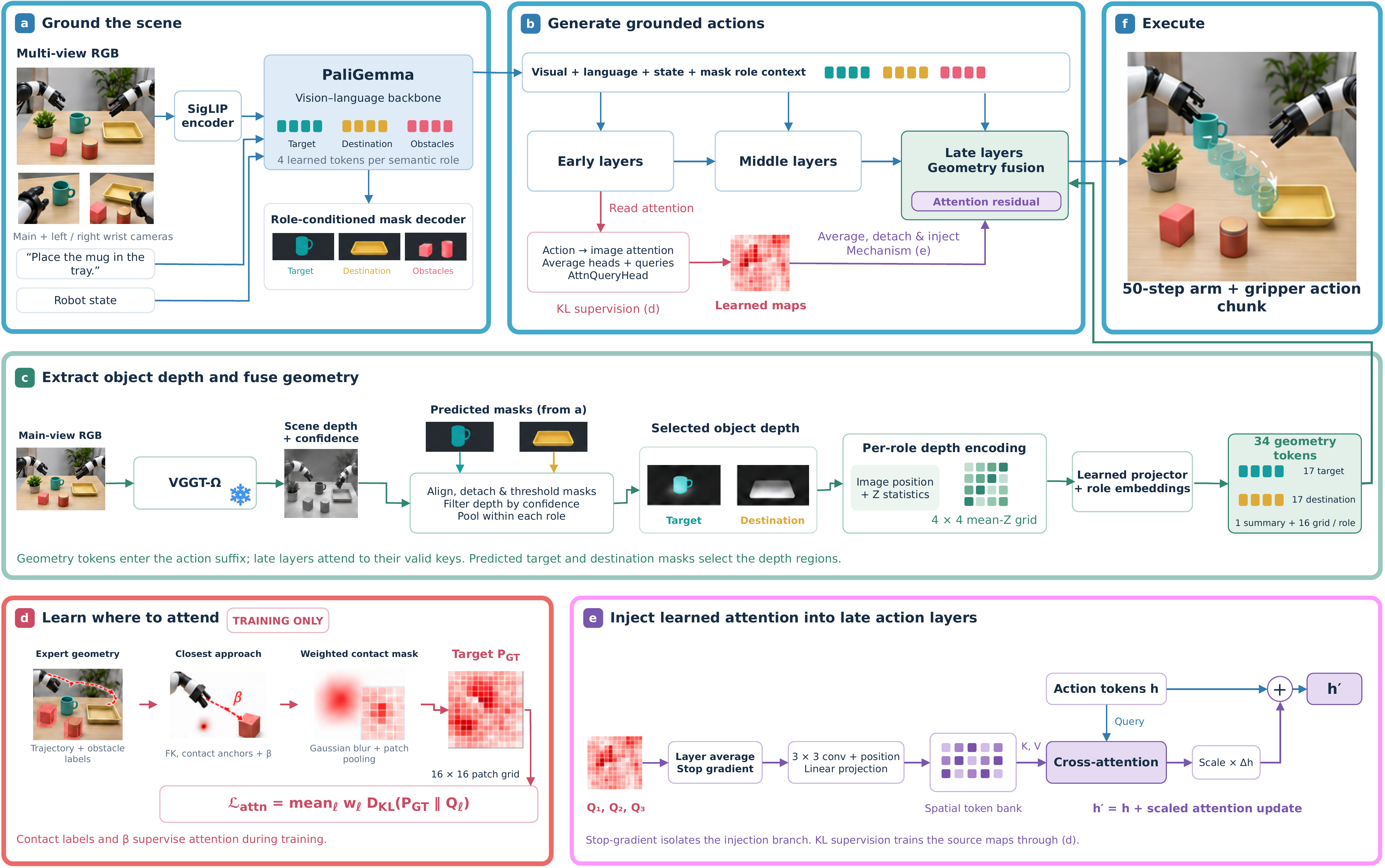}
    \caption{\textbf{MOSAIC Model Structure.} (a) A vision–language backbone encodes observations, instructions, and robot state, with role tokens supporting semantic mask prediction. (b) The action expert attends to the shared prefix and produces learned attention maps. (c) Object-level depth is extracted using the VGGT-$\Omega$ combined with the predicted masks. The depth tokens are generated and injected into the action expert final layers. (d) Contact-relevant expert targets supervise attention through forward KL divergence during training. (e) Predicted attention maps are averaged, detached, and injected through residual cross-attention into late action layers. (f) Flow matching generates actions for robot execution.}
    \label{fig:main} \vspace{-4mm}
\end{figure*}

\section{Methodology}


Our motivation is to develop modular components that enhance the object-level scene understanding of VLA models and improve their performance in cluttered environments, which is detailed in Fig.~\ref{fig:main}. We first describe the general VLA architecture in Section~\ref{sec:vla}. Then in Section~\ref{sec:masks}, we introduce our semantic mask generation module and explain how the action expert is conditioned on the resulting representations implicitly. In Section~\ref{sec:inject}, we describe how object-level depth information is injected into the action expert. Finally, Section~\ref{sec:attn} presents our attention-guidance strategy, which directs the action expert toward both the target object and surrounding obstacles.

\subsection{Preliminaries: Vision-Language-Action Model} \label{sec:vla}

A shared vision--language backbone encodes the visual observation and language instruction into a contextual prefix, while a smaller action expert generates a horizon of robot actions as a suffix conditioned on these features. Information flows from the backbone to the action expert through attention, while the backbone representations remain independent of the generated action tokens.
Given a multi-view observation \(I=\{I_v\}_{v=1}^{V}\) and a language
instruction \(\ell\) (with proprioceptive state embedded as discrete
tokens), the vision encoder and language tower produce a
sequence.
Suffix attention is causal over the action tokens and is allowed to
read the entire prefix through a cross-attention block.
A joint transformer step then yields:
\vspace{-2mm}
\begin{equation}
  \bigl[\mathbf{h}^{\mathrm{pre}},\;\mathbf{h}^{\mathrm{suf}}\bigr]
  =
  \mathrm{Transformer}\!\left(
    \bigl[\mathbf{z}^{\mathrm{pre}},\;\mathbf{z}^{\mathrm{suf}}\bigr],\;
    \mathcal{M}
  \right)
\end{equation}
where $\mathbf{z}^{\mathrm{pre}}$ contains the visual and language tokens, while $\mathbf{z}^{\mathrm{suf}}$ is obtained from the noisy action chunk. The attention mask $\mathcal{M}$ allows action tokens to attend to the contextualized vision--language features while preserving causal attention among actions. The resulting $\mathbf{h}^{\mathrm{pre}}$ and $\mathbf{h}^{\mathrm{suf}}$ denote the corresponding hidden representations ($\mathbf{h}$) with only $\mathbf{h}^{\mathrm{suf}}$ decoded into an action-space velocity.

\vspace{-3mm}
\begin{equation}
  \mathbf{v}_t
  =
  W_{\mathrm{out}}\,\mathbf{h}^{\mathrm{suf}}
  \in \mathbb{R}^{H\times A}
\end{equation}

where \(\mathbf{v}_t\) is the instantaneous velocity of the action chunk in flow space. 

\vspace{-2mm}
\subsection{Semantic Mask Generation for Objects} \label{sec:masks}
To improve the VLA model's understanding of its environment, we propose auxiliary semantic mask prediction task within the VLM backbone.
As Fig.~\ref{fig:main}(a) shows, the masks are divided into three task-specific roles: the target object, destination region, and surrounding obstacles. This role-specific representation helps the model preserve their distinct semantics and use them appropriately during action generation. Given the visual observation and the contextual information extracted from the language instruction, the backbone learns to predict these semantic regions jointly with its standard representations.
We reserve a dedicated token group for each role, \(e^r\), containing \(K\) tokens, and append it to the VLM prefix. 
\begin{equation}
\begin{aligned}
  \mathbf{e}^{r}
  &\in
  \mathbb{R}^{K\times d},
  \qquad
  r\in\{\mathrm{tar},\mathrm{dst},\mathrm{obs}\} \\
  \mathbf{z}^{\mathrm{pre}}
  &=
  \bigl[
    \phi(I),\;
    \tau(\ell),\;
    \mathbf{e}^{\mathrm{tar}},\;
    \mathbf{e}^{\mathrm{dst}},\;
    \mathbf{e}^{\mathrm{obs}}
  \bigr]
\end{aligned}
\end{equation}
where \(\phi\) denotes the visual encoder's image tokens and \(\tau\) the tokenized prompt. These role tokens participate in bidirectional prefix attention together with the visual and language tokens, therefore each group encodes task-conditioned visual context for its semantic role rather than a generic objectness signal.
Following the shared vision--language forward pass, a dedicated mask head uses the role tokens to attend to the primary-camera patch grid.
The \(K\) tokens of role \(r\) are then mean-pooled into a single query, and the three queries ($\mathbf{q}$) independently decode target, destination, and obstacle masks over the same image.
\vspace{-2mm}
\begin{equation}
\resizebox{0.98\linewidth}{!}{$
  \bigl[M^{\mathrm{tar}}, M^{\mathrm{dst}}, M^{\mathrm{obs}}\bigr]
  =
  \mathrm{MaskHead}\!\left(
    \mathbf{q}^{\mathrm{tar}},\!
    \mathbf{q}^{\mathrm{dst}},\!
    \mathbf{q}^{\mathrm{obs}},\!
    \phi(I_{\mathrm{prim}})
  \right).
$}
\end{equation}

A coarse semantic map is first predicted from the visual patch grid and then refined with a shallow RGB residual branch to preserve fine object boundaries. Each valid semantic role is supervised independently using weighted binary cross-entropy, soft Dice, and boundary losses, with the final objective averaged only over valid roles. This avoids penalizing tasks in which a destination or obstacle is absent. The resulting masks preserve clear role distinctions between the target object, destination region, and surrounding obstacles while maintaining accurate boundaries. During action generation, as Fig.~\ref{fig:main}(b) demonstrates, each suffix action token attends to the complete VLM prefix, including the role-specific mask tokens:

\vspace{-4mm}

\begin{equation}
  \mathbf{h}^{\mathrm{suf}}_{i}
  =
  \mathrm{Attn}\!\left(
    \mathbf{z}^{\mathrm{suf}}_{i},\;
    \bigl[
      \mathbf{h}^{\mathrm{vis}},\;
      \mathbf{h}^{\mathrm{lang}},\;
      \mathbf{h}^{\mathrm{tar}},\;
      \mathbf{h}^{\mathrm{dst}},\;
      \mathbf{h}^{\mathrm{obs}}
    \bigr]
  \right).
\end{equation}
This transfers the disentangled grounding from the VLM backbone to the action expert, so the predicted flow velocity \(\mathbf{v}_t\) and therefore the executed action chunk can explicitly reason about the object to manipulate, its destination, and the surrounding obstacles.

\subsection{Object Depth Injection into Action Expert} \label{sec:inject}

To improve the action expert's understanding of task-relevant objects, as in Fig.~\ref{fig:main}(c), we extract object-level depth information and incorporate it directly into the action-generation process. Specifically, we provide metric depth cues for the target object and destination region according to their semantic roles. A frozen 3D scene reconstruction decoder, \(\Omega\), maps the primary-camera image to a scene depth map and confidence map, \((D,C)=\Omega(I_{\mathrm{prim}})\).
Then the backbone mask head predicts the role-specific masks \(M^{r}\). For each role, only pixels with valid depth range and sufficient confidence are retained. The resulting geometric information is processed independently for each semantic role rather than being merged into a single point cloud. Using these masks, we construct two complementary geometric encodings. The first is a compact statistics token:

\vspace{-2mm}
\begin{equation}
  \mathbf{s}^{r}
  =
  \bigl[
    \bar{u}^{r},\;
    \bar{v}^{r},\;
    u^{r}_{0},v^{r}_{0},u^{r}_{1},v^{r}_{1},\;
    \bar{z}^{r},\;
    z^{r}_{\min},\;
    \sigma^{r}_{z}
  \bigr].
\end{equation}

The statistics token captures the masked image centroid, $u^r, v^r$, the bounding region, $u^{r}_{0},v^{r}_{0},u^{r}_{1},v^{r}_{1}$, and the mean, minimum, and standard deviation of the depth, $\bar{z}^{r},\; z^{r}_{\min},\; \sigma^{r}_{z}$. While this representation provides the global image location and depth distribution of each semantic role, it does not preserve local geometric variations within the region.

To retain this spatial information, we introduce a second encoding by partitioning the same region into a $G\times G$ grid (G=4). Each grid cell is represented by its center $(u^{r}_{ij},v^{r}_{ij})$ and mean metric depth $\bar{z}^{r}_{ij}$. A shared projection module composed of LayerNorm, GELU, and a linear layer maps both the global statistics vector $\mathbf{s}^{r}$ and the $(G^{2})$ local grid features to the action-expert embedding dimension. Finally, a learned role embedding $\mathbf{p}^{r}$ is added to each token, allowing the action expert to distinguish between target and destination representations even when their geometric features are similar. This results in $(N_r = 1 + G^2 = 17)$ geometry tokens for each semantic role, forming
$\mathbf{z}^{\mathrm{geo}} \in \mathbb{R}^{|\mathcal{R}|N_r \times d_{\mathrm{act}}}$
where $(\mathcal{R}={\mathrm{tar},\mathrm{dst}})$. These geometry tokens are then added to the noisy action-token sequence before being processed by the action expert. 
\vspace{-3mm}
\begin{equation}
  \mathbf{z}^{\mathrm{suf}}
  =
  \bigl[\mathbf{z}^{\mathrm{geo}},\;W_{\mathrm{in}}\mathbf{x}_{t}\bigr].
\end{equation}

Action queries attend to \(\mathbf{z}^{\mathrm{geo}}\) only in the final six layers of the action expert. This allows the earlier layers to primarily reason from the VLM prefix, including the semantic mask tokens, while the final layers refine the predicted actions using explicit metric geometry.
We found that the same representation is not suitable for surrounding obstacles. Therefore, obstacle information is incorporated using the attention-guidance strategy described in Section~\ref{sec:attn}. 

\subsection{Attention Guidance on the Action Expert} \label{sec:attn}
\begin{figure*}\vspace{2mm}
    \centering
    \begin{subfigure}[c]{0.16\textwidth}
        \centering %
        \includegraphics[height=2.3cm, keepaspectratio]{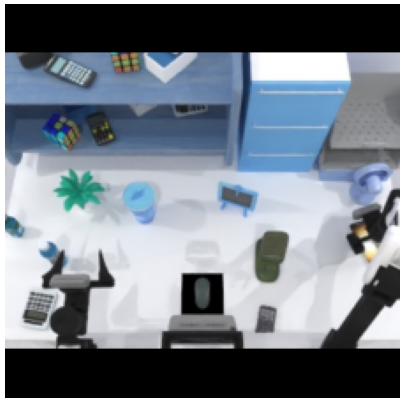}
        \caption{}
        \label{fig:rgb}
    \end{subfigure}
    \hfill
    \begin{subfigure}[c]{0.16\textwidth}
        \centering %
        \includegraphics[height=2.3cm, keepaspectratio]{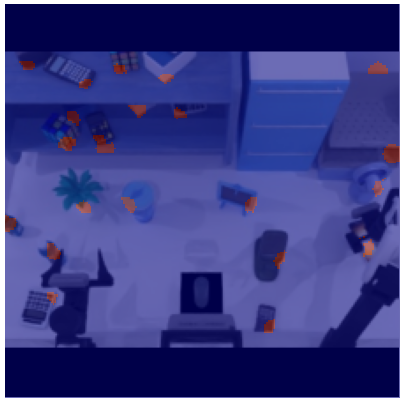}
        \caption{}
        \label{fig:contact_disks}
    \end{subfigure}
    \hfill
    \begin{subfigure}[c]{0.16\textwidth}
        \centering %
        \includegraphics[height=2.3cm, keepaspectratio]{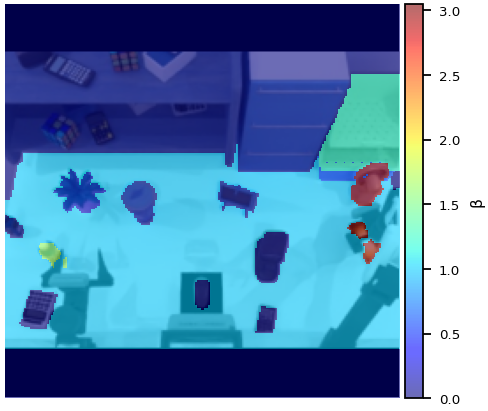}
        \caption{}
        \label{fig:contact_beta}
    \end{subfigure}
    \hfill
    \begin{subfigure}[c]{0.16\textwidth}
        \centering %
        \includegraphics[height=2.3cm, keepaspectratio]{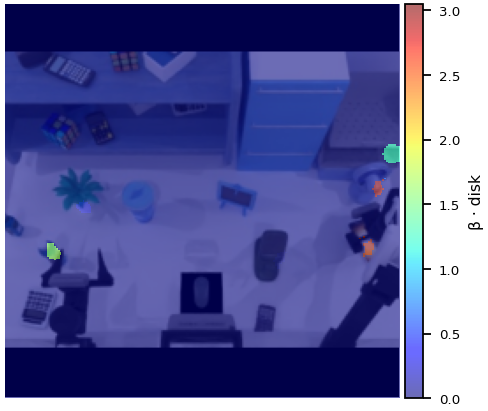}
        \caption{}
        \label{fig:obstacle_beta}
    \end{subfigure}
    \hfill
    \begin{subfigure}[c]{0.16\textwidth}
        \centering %
        \includegraphics[height=2.3cm, keepaspectratio]{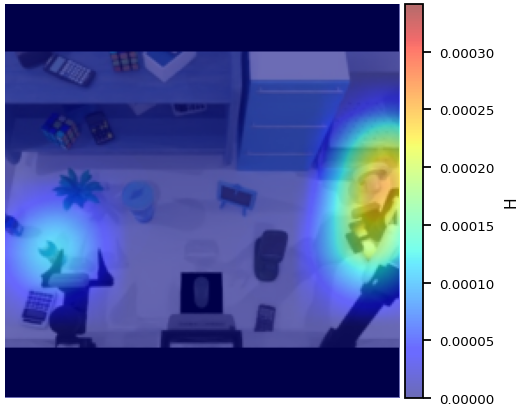}
        \caption{}
        \label{fig:heatmap}
    \end{subfigure}
    \hfill
    \begin{subfigure}[c]{0.16\textwidth}
        \centering %
        \includegraphics[height=2.3cm, keepaspectratio]{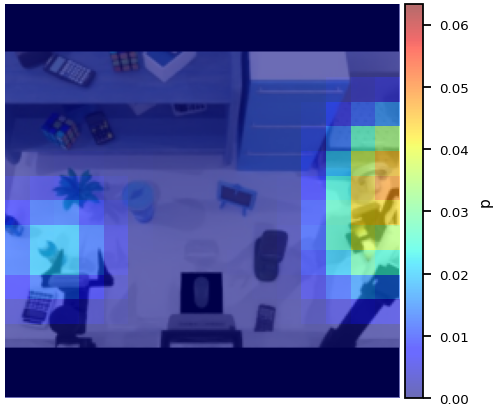}
        \caption{}
        \label{fig:attention}
    \end{subfigure}
    \caption{
        The pipeline for groundtruth attention map generation.
        \textbf{(a)} The original RGB countertop camera view. 
        \textbf{(b)} Per-object contact disk generation. The closest point of each object's bounding box is projected onto pixel space. Disks are centered at these points to highlight potential contact zones.
        \textbf{(c)} Proximity weight map $\beta$. Higher weight for objects closer to the robot body, less weight for farther objects, zero weight for very distant objects.
        \textbf{(d)} Obstacle contact mask. Using the proximity weight map $\beta$ to filter out irrelevant objects.
        \textbf{(e)} Groundtruth attention heatmap $\mathbf{H}$. Obstacle contact mask convolved with a Gaussian kernel with standard deviation $\sigma$ to represent the spatial distribution of visual attention. 
        \textbf{(f)} Patch-wise distributed groundtruth attention map. Matches the granularity of the action expert's attention modules.
    }
    \label{fig:obs-pipeline} \vspace{-4mm}
\end{figure*}

To improve awareness of task-relevant regions, we incorporate contact-aware object attention supervision to teach the model to look at objects in the environment that lead to unsafe scenarios, shown in Fig.~\ref{fig:main}(d). For a given object $k$ defined by its object-oriented bounding box (OOBB) center $c_k$, half-extents $h_k$, and rotation matrix from OOBB to world coordinates $R_k$, we first convert a robot point $p$ to object $k$'s coordinate frame and compute its distance to the OOBB:
\vspace{-2mm}
\begin{equation}\label{eq:attn1}
\resizebox{0.9\linewidth}{!}{%
    $x = R_k^\top(p-c_k) \in \mathbb{R}^3, \, d_{\mathrm{OOBB}}(p, k) = ||\text{max}(0, \text{abs}(x) - h_k)||_2$
    }
\end{equation}

\noindent where $\text{abs}(\cdot)$ is a component-wise absolute value operation on a vector. Note that if $p$ lies within the object's OOBB, then $d(p, k) = 0$. This distance is then calculated per frame by 
\vspace{-2mm}
\begin{equation}
    d(\tau, k) = \min_i d_{\mathrm{OOBB}}(p_i(\tau), k) \label{disperframe}
\end{equation}
where $i$ is the index of the robot point and $\tau$ denotes the timestep. The closest surface point is stored for object tracking. To incorporate temporal understanding and allow the policy to anticipate potential collisions, the minimum distance is calculated over a temporal window of width $W$ centered at timestep $t$ by 
\begin{equation}
    d_{\mathrm{min}}(t, k) = \min_{\tau \in [t-W, t+W]} d(\tau, k). \label{mindistperframe}
\end{equation}

The contact point at timestep $t$ is the same surface point that corresponds to the minimizer $\tau^*$. Then, for a user specified scale $g$, threshold $d_{\mathrm{thresh}}$, and soft threshold $m$, the proximity weights are calculated by 
\vspace{-1mm}
\begin{equation}
\resizebox{0.98\linewidth}{!}{%
    $\beta(t, k) = \begin{cases}
        1 + g \left(1 - \frac{d_{\mathrm{min}}(t,k)}{d_{\mathrm{thresh}}}\right) &d_{\mathrm{min}}(t,k) < d_{\mathrm{thresh}} \\
        1 - \text{clip}\left(\frac{d_{\mathrm{min}}(t,k) - d_{\mathrm{thresh}}}{m}, 0, 1\right) &d_{\mathrm{min}}(t,k) \geq d_{\mathrm{thresh}} 
    \end{cases} \label{betaweight}$
  }
\end{equation}
\vspace{-1mm}
where the $\text{clip}$ operator clamps values between 0 and 1.

The closest points of each object to the robot are highlighted, creating a per-object contact-aware mask. These masks are aggregated by a pixel-wise max operation over all objects to form $\Tilde{\mathbf{m}}_t$. These aggregated masks follow a similar procedure to that of Gaze-VLM \cite{pani2026gazevlm} for generating spatial heatmaps $\mathbf{m}_t$ and converting it to a patch-wise groundtruth attention distribution $\Tilde{\text{H}}_t$. 

This process provides groundtruth attention maps for obstacles and, optionally, target objects and destination regions. The obstacle contact mask is multiplied pixel-wise by the aggregated proximity weight map $\beta_t$ prior to spatial heatmap generation, i.e. $\Tilde{\mathbf{m}}_t \odot \beta_t$, where $\beta_t$ is initialized as all ones and entry $\beta_t[u,v]$ is updated to $\beta_{t,k}$ if object $k$'s segmentation mask occupies pixel $(u,v)$.

For each supervised attention layer, the $N_\text{actions} \times N_\text{vis}$ attention block where the action tokens attend to the visual tokens is extracted, renormalized, and averaged over all non-batch and non-visual token dimensions to allow for specialization. The resulting block is resized and passes through a dedicated AttentionQueryHead, which learns a residual on the attention values in log-space. 

The groundtruth attention maps $\Tilde{\text{H}}_t$ can be compared with the AttentionQueryHead output $A_t$. A KL-divergence term is used to encourage the model to focus on obstacles, and optionally the target object and destination region:
\vspace{-2mm}
\begin{equation} \label{eq:attn4}
    D_{\mathrm{KL}} \left(\Tilde{\text{H}}_t || A_t \right) = \sum_{i=1}^P \Tilde{\text{H}}_{t, i} \log \frac{\Tilde{\text{H}}_{t, i}}{A_{t, i}}
\end{equation}
\vspace{-3mm}

\noindent where $\Tilde{\text{H}}_{t, i}, A_{t,i}$ corresponds to the $i$-th patch of the corresponding attention map. Each supervised layer and obstacle region has their own KL loss term and scaled by a loss weight. The average over all supervised layers and regions is added to the overall training objective. As Fig.~\ref{fig:main}(e) shows, the output of the AttentionQueryHead is also injected into later layers through a cross-attention-based residual stream on the action tokens to provide more task-aware context for action generation. Fig.~\ref{fig:obs-pipeline} visually demonstrates the steps for the attention guidance map generation.

\section{Experiments}

\begin{figure*}[t]
    \centering
    \includegraphics[width=\textwidth]{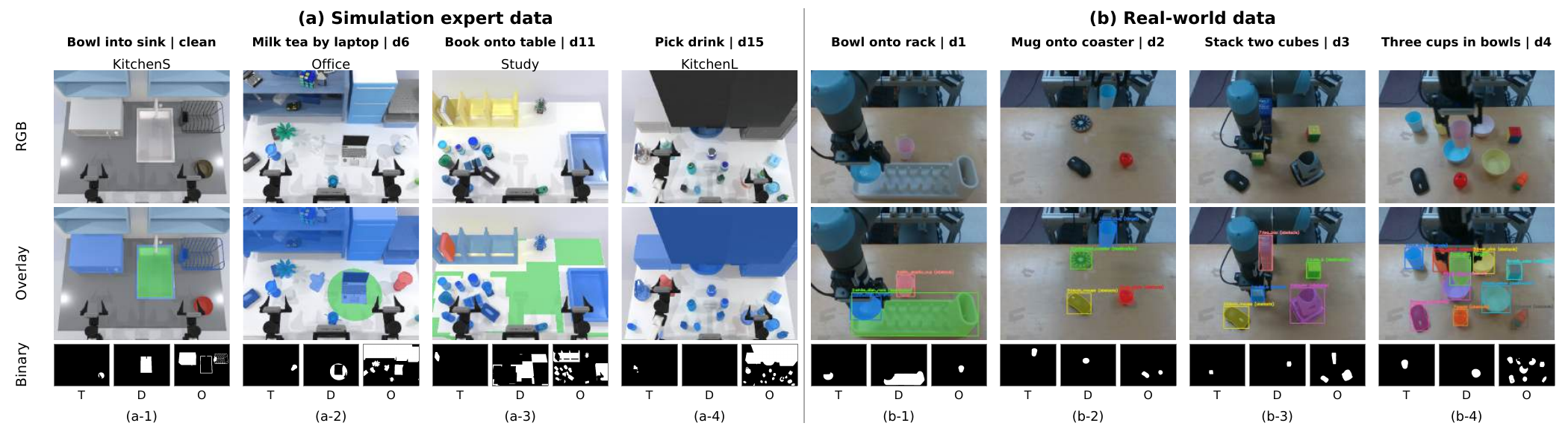}
    \caption{Task-conditioned visual annotations for (a) Simulation and (b) Real-world data. In (a), rows show RGB images, mask overlays, and binary target (T), destination (D), and obstacle (O) masks. Examples illustrate an object destination \mbox{(a-1)}, anchor-relative placement \mbox{(a-2)}, a table-region destination with 5 cm clearance \mbox{(a-3)}, and pick-only manipulation without a destination \mbox{(a-4)}. Simulation overlays use red/green/blue for T/D/O. In (b), examples (b-1) to (b-4) show four real-world tasks with one to four added distractors.}
    \label{fig:data-grounding} \vspace{-3mm}
\end{figure*}

\vspace{-2mm}
\subsection{Data Generation}

\subsubsection{Simulation}
\label{sec:simulation-data}
We use RoboPRO~\cite{li2026robopro} as our primary simulation benchmark because it targets manipulation in cluttered environments, a setting that most existing robotic manipulation benchmarks do not emphasize. RoboPRO consists of four furnished
environments, namely Office, Study, Kitchen-S, and Kitchen-L, containing 80 tasks across three categories: single-object,
articulated-object, and multi-step. Each task provides 200 expert demonstrations: 100
collected in clean scenes containing only the target objects and basic furniture, and
100 collected in cluttered scenes populated with additional distractors. The number of distractors ranges from 6 to 15 per scene. RoboPRO also provides tools for applying visual, linguistic, and object-level perturbations at run time in both clean and cluttered scenes.

We replay feasible seeds that pass task-specific success checks for data capture. Each episode records synchronized $320\times240$ RGB-D observations from the countertop and both wrist cameras, along with camera intrinsics and extrinsics, end-effector poses, and 14-dimensional joint-target and gripper commands. Instance IDs, object poses, and oriented bounding boxes for rigid actors and articulated links preserve pixel-to-object correspondence for geometric annotation.

We derive semantic roles automatically from object motion, gripper proximity, and task metadata. For each manipulation stage, the manipulated object is assigned as the target, while its destination is inferred from the final pose, container geometry, articulated-link bounds, and task intent. Surface and next-to placements use projected geometric regions, whereas pick-only and opening/closing actions may have no destination. Roles are updated across stages in multi-step tasks. After defining the target and destination masks, the obstacle masks are constructed from the remaining visible instances, excluding the robot and scene. The robot and objects' 3D geometry and robot trajectories further support proximity-based attention supervision (Section~\ref{sec:attn}), providing pre-contact cues in addition to measured collisions. Fig.~\ref{fig:data-grounding}(a) shows examples from all four environments.

\subsubsection{Real-world}
\label{sec:real-data}
For real-world tests, we collect 1,000 annotated episodes using a UR5e arm (Universal Robots) with a Robotiq 2F-85 parallel gripper. The dataset covers five tasks: placing a cup in a bowl, a bowl on a rack, a mug on a coaster, two-cube stacking, and sequentially placing three cups into three bowls as a compositional task. Each task includes 100 clean episodes and 25 episodes at each clutter level $(d1-d4)$, corresponding to one to four additional distractors.
Three Intel RealSense cameras capture base, wrist, and fixed countertop views, while only the countertop camera is used for model input and annotation. RGB and aligned depth are recorded at $640\times480$ resolution at approximately 20 Hz, with timestamps retained. We also record joint states, gripper positions, camera intrinsics, and depth scale for 3D reconstruction.
Persistent instance masks are generated using prompted SAM~2.1 video segmentation~\cite{ravi2024sam2}, followed by visual review and corrections. Each instance is assigned a target, destination, or obstacle role, with modal 2D bounding boxes derived from the visible masks. The robot and unannotated background are excluded from obstacle labels. Fig.~\ref{fig:data-grounding}(b) presents examples from four tasks, showing the three-cup compositional task in place of the single-cup placement task.
\vspace{-2mm}

\subsection{Implementation Details}
We implement MOSAIC on top of $\pi_{0.5}$, one of the most widely used open-source VLA models. 
We first fully fine-tune and converge the base $\pi_{0.5}$ models on the RoboPRO data for 30K steps with a batch size of 192, using only images and language instructions as inputs and robot actions as outputs to adapt the policy to RoboPRO. This checkpoint is the baseline in Tables~\ref{tab:pi05_ablation} and~\ref{tab:vision_perturbation}.
From this base checkpoints, we train MOSAIC in two stages across all four RoboPRO environments with a global batch size of 64. In Stage 1, we train the semantic grounding module for one epoch using 12 learnable mask token slots, with 4 tokens assigned to each of the target, destination, and obstacle roles. We optimize the grounding heads and rank-32 LoRA~\cite{hu2021lora} adapters applied to the vision encoder and language backbone, while keeping the original backbone and action expert frozen. The LoRA weights are used in this stage in order to keep the original knowledge of the VLA's backbone for the action generation. The grounding heads and LoRA adapters use peak learning rates of $10^{-4}$ and $5\times10^{-6}$, respectively, with 10\% warmup followed by cosine decay to one-tenth of their peak values.

In Stage 2, we merge the LoRA weights into the backbone and jointly fine-tune the full policy for two epochs using flow-matching action prediction, mask supervision, and obstacle-attention supervision. Depth features from a frozen VGGT-$\Omega$ model are pooled using the predicted target and destination masks and injected into the final six action-expert layers. In parallel, our attention-guidance module supervises the first three action-expert layers and injects the resulting attention features into the final three layers to further guide action generation. The learning rate peaks at $2.5\times10^{-5}$ after a 5\% warm-up and decays to $2.5\times10^{-6}$, with a $10\times$ learning-rate multiplier applied to the attention-injection parameters. Both stages use AdamW with $(\beta_1,\beta_2)=(0.9,0.95)$, $\epsilon=10^{-8}$, weight decay $10^{-10}$, and global gradient clipping at 1.0. Stage 2 also uses an exponential moving average with decay 0.99. The policy takes three $224\times224$ camera views as input and predicts 50-step action chunks. For attention guidance, we set $W = 7$ timesteps, $d_{\mathrm{thresh}} = 15$\,cm, $m = 5$\,cm, $g = 3$, and the Gaussian kernel width to $\sigma = 15$\,px.

\subsection{Results on RoboPRO}

Table~\ref{tab:pi05_ablation} reports per-environment success rate (SR), hard success rate (HSR), and collision rate (CR) under clean and cluttered settings. HSR denotes episodes that complete the task successfully without any unintended collision. Averaged across all four environments, adding masks, attention, and depth successively increases overall SR. Masks alone improve SR over the $\pi_{0.5}$ (RoboPRO) baseline by $1.0\%$ in clean scenes and $2.7\%$ in cluttered scenes. Also, removing obstacle-mask conditioning lowers SR in cluttered scenes from $55.4\%$ to $52.7\%$ and HSR from $39.2\%$ to $38.7\%$. This indicates that explicit obstacle representations provide complementary context beyond target- and destination-only conditioning.
Attention module adds $0.5\%$ and $1.5\%$, while object-level depth injection increases the total gain by $+4.1\%$ SR in clean scenes and $+5.1\%$ SR in cluttered scenes. The largest gains occur in challenging cluttered settings: in Office, our full model improves SR by $9.9\%$ and HSR by $7.1\%$, while in Kitchen-S, SR/HSR improves by $4.2$/$1.9\%$.

In clean scenes, HSR closely follows SR ($61.0\% \rightarrow 64.8\%$), indicating that most additional successes remain collision-free. In cluttered scenes, however, the HSR improvement is smaller ($39.7\% \rightarrow 41.3\%$), while the average CR increases from $39.7\%$ to $44.8\%$. This suggests a behavioral trade-off: the grounded policy attempts and completes more manipulations in cluttered workspaces, increasing both successful interactions and exposure to potential collisions. 
These results suggest that explicit grounding improves the policy's ability to act in clutter, while further improvements in execution safety remain helpful.

\begin{table*}[t]\vspace{1mm}
\centering
\small
\setlength{\tabcolsep}{2.4pt}
\caption{Per-environment ablation results for $\pi_{0.5}$ on RoboPRO. We evaluate the contribution of each introduced module. Best per column within each scene in \textbf{bold}, second best \underline{underlined}. All variants are evaluated on the same episodes.}\vspace{-2mm}

\begin{tabular}{ll ccc | ccc ccc ccc ccc}
\toprule
& & \multicolumn{3}{c}{\textbf{All}} & \multicolumn{3}{c}{Kitchen-L} & \multicolumn{3}{c}{Kitchen-S} & \multicolumn{3}{c}{Office} & \multicolumn{3}{c}{Study} \\
\cmidrule(lr){3-5} \cmidrule(lr){6-8} \cmidrule(lr){9-11} \cmidrule(lr){12-14} \cmidrule(lr){15-17}
Scene & $\pi_{0.5}$ variant & SR$\uparrow$ & HSR$\uparrow$ & CR$\downarrow$ & SR$\uparrow$ & HSR$\uparrow$ & CR$\downarrow$ & SR$\uparrow$ & HSR$\uparrow$ & CR$\downarrow$ & SR$\uparrow$ & HSR$\uparrow$ & CR$\downarrow$ & SR$\uparrow$ & HSR$\uparrow$ & CR$\downarrow$ \\
\midrule

\multirow{5}{*}{Clean}
& RoboPRO
& 66.0 & 61.0 & \textbf{11.9}
& 65.8 & 56.0 & \textbf{21.2}
& 68.1 & 61.5 & \textbf{12.4}
& 54.8 & 53.2 & \textbf{8.5}
& 75.9 & 73.6 & \textbf{5.4} \\

& MOSAIC (Masks w/o Obst.)
& \underline{67.8} & 62.1 & 16.0
& \underline{69.0} & 56.0 & 27.0
& 69.6 & 63.6 & 13.0
& 54.2 & 53.2 & \underline{14.5}
& \textbf{79.0} & \textbf{76.2} & 9.0 \\

& MOSAIC (Masks)
& 67.0 & 61.8 & 16.3
& 68.2 & \underline{57.2} & 26.0
& 67.5 & 62.0 & 14.8
& \textbf{60.2} & \textbf{58.0} & 16.5
& 72.1 & 70.0 & 7.4 \\

& MOSAIC (Masks+Attn)
& 67.5 & \underline{62.2} & 15.6
& 68.5 & \underline{57.2} & 26.5
& \underline{71.0} & \underline{64.4} & 14.8
& 53.0 & 51.8 & 15.0
& \underline{77.9} & \underline{75.9} & \underline{5.9} \\

& MOSAIC (Masks+Attn+Depth)
& \textbf{70.1} & \textbf{64.8} & \underline{14.5}
& \textbf{74.0} & \textbf{61.8} & \underline{23.0}
& \textbf{71.8} & \textbf{66.0} & \underline{12.9}
& \underline{58.2} & \underline{57.0} & 16.2
& 76.7 & 74.9 & \textbf{5.4} \\

\midrule

\multirow{5}{*}{Cluttered}
& RoboPRO
& 52.7 & 39.7 & \textbf{39.7}
& \underline{57.0} & \textbf{34.5} & \textbf{55.2}
& \underline{59.6} & \underline{47.1} & \textbf{28.6}
& 36.2 & 30.9 & \textbf{41.8}
& 58.0 & 46.5 & \underline{32.7} \\

& MOSAIC (Masks w/o Obst.)
& 52.7 & 38.7 & \underline{43.1}
& 55.8 & \underline{29.9} & \underline{63.3}
& 54.7 & 43.2 & 29.9
& 39.1 & 30.2 & 45.7
& 61.4 & \underline{51.8} & 32.8 \\

& MOSAIC (Masks)
& 55.4 & 39.2 & 44.1
& \underline{57.0} & 28.5 & 64.8
& 57.6 & 45.3 & \underline{29.7}
& 44.1 & 32.2 & 47.1
& 63.1 & 51.0 & 34.4 \\

& MOSAIC (Masks+Attn)
& \underline{56.9} & \underline{40.4} & 44.3
& 55.2 & 26.5 & 66.8
& 58.9 & 44.0 & 32.8
& \textbf{46.6} & \underline{35.2} & 44.8
& \textbf{66.8} & \textbf{56.0} & \textbf{32.4} \\

& MOSAIC (Masks+Attn+Depth)
& \textbf{57.8} & \textbf{41.3} & 44.8
& \textbf{58.0} & 27.0 & 64.5
& \textbf{63.8} & \textbf{49.0} & 31.8
& \underline{46.1} & \textbf{38.0} & \underline{43.8}
& \underline{63.6} & \underline{51.8} & 38.4 \\

\bottomrule
\end{tabular}
\label{tab:pi05_ablation}
\end{table*}

\begin{figure*}
    \centering
    \begin{subfigure}[c]{0.49\textwidth}
        \centering %
        \includegraphics[width=\textwidth]{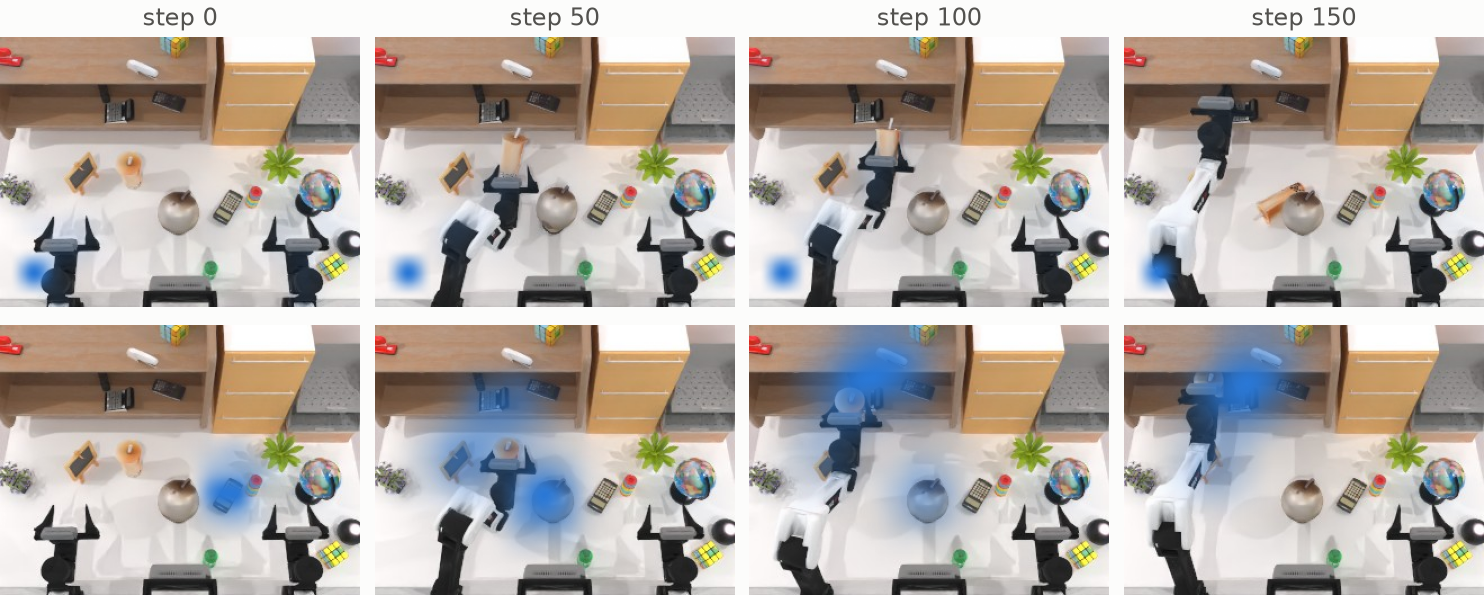}
        \label{fig:attn1}
    \end{subfigure}
    \hfill
    \begin{subfigure}[c]{0.49\textwidth}
        \centering %
        \includegraphics[width=\textwidth]{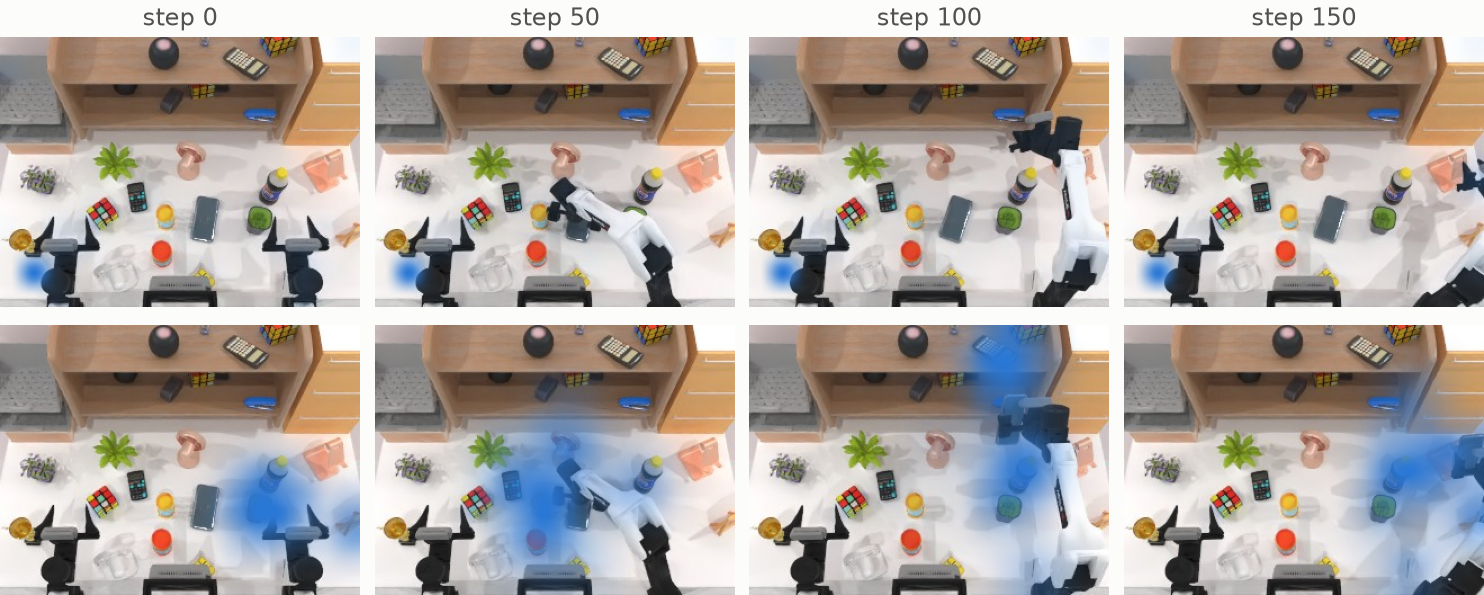}
        \label{fig:attn2}
    \end{subfigure} \vspace{-6mm}
    \caption{
    Average attention maps from the first three action-expert layers for the RoboPRO-finetuned base $\pi_{0.5}$ model (top) and our final model (bottom) across two manipulation actions. Blue regions indicate higher attention. Our model consistently focuses on potential collision regions, while the base $\pi_{0.5}$ exhibits less obstacle-aware attention.}
    
    
    \label{fig:attn-comparison} 
\end{figure*}

 To measure robustness to visual changes that preserve the task and scene geometry, we evaluate our policy against the base $\pi_{0.5}$ (RoboPRO) policy under RoboPRO's visual perturbation setting. Each episode is built from the same seed as its clean counterpart, so object layout, initial robot state and instruction are identical to the clean evaluation and only the rendering and the camera images change. Three perturbations are applied together. (i) Lighting: each directional and point light receives an RGB tint, sampled uniformly from $[0.4, 1.8]$, the directional light is rotated to a random elevation in $[8^\circ, 82^\circ]$ and a random azimuth, and shadow casting is switched off with probability $0.5$. (ii) Blur: every RGB observation is Gaussian-blurred with $\sigma = 0.5$ px. (iii) Pixel shift: every RGB observation is translated by an offset drawn uniformly from $[-0.5, 0.5]$ px on each axis, independently per frame and per camera, with reflected borders.

We evaluate 12 RoboPRO perturbation tasks, with three default tasks each in Office, Study, Kitchen-L, and Kitchen-S. Each task is evaluated on 20 fixed clean benchmark seeds with seen language instructions and a fixed policy seed. We report SR, HSR, and CR in Table~\ref{tab:vision_perturbation}. Our method consistently improves robustness, increasing overall SR from $65.2\%$ to $74.9\%$ ($+9.7\%$) and HSR from $63.4\%$ to $72.7\%$ ($+9.3\%$), while keeping CR nearly unchanged. The largest gain occurs in Office, where SR improves from $51.7\%$ to $76.7\%$ ($+25.0\%$) and HSR from $50.0\%$ to $71.7\%$ ($+21.7\%$). The HSR gain is close to the SR gain, indicating that the recovered successes remain largely collision-free. These results suggest that our modules provide task-relevant spatial structure that remains robust to changes in visual appearance.
Fig.~\ref{fig:attn-comparison} provides further qualitative evidence. Our model consistently attends to potential collision regions throughout the rollout, whereas the base $\pi_{0.5}$ (RoboPRO) model often allocates attention to less relevant regions. The examples show greater attention to potential contact regions, alongside improved robustness to visual perturbations. 
\begin{table*}[t]\vspace{1mm}
\centering
\small
\setlength{\tabcolsep}{3.5pt}
\caption{Robustness to visual perturbation (lighting, blur and per-frame pixel shift) on clean RoboPRO scenes. Best results within each model pair are shown in \textbf{bold}, including ties.}\vspace{-2mm}
\label{tab:vision_perturbation}
\begin{tabular}{l ccc | ccc ccc ccc ccc}
\toprule
& \multicolumn{3}{c}{\textbf{All}} & \multicolumn{3}{c}{Kitchen-L} & \multicolumn{3}{c}{Kitchen-S} & \multicolumn{3}{c}{Office} & \multicolumn{3}{c}{Study} \\
\cmidrule(lr){2-4} \cmidrule(lr){5-7} \cmidrule(lr){8-10} \cmidrule(lr){11-13} \cmidrule(lr){14-16}
Model & SR$\uparrow$ & HSR$\uparrow$ & CR$\downarrow$ & SR$\uparrow$ & HSR$\uparrow$ & CR$\downarrow$ & SR$\uparrow$ & HSR$\uparrow$ & CR$\downarrow$ & SR$\uparrow$ & HSR$\uparrow$ & CR$\downarrow$ & SR$\uparrow$ & HSR$\uparrow$ & CR$\downarrow$ \\
\midrule







$\pi_{0.5}$ (RoboPRO)
& 65.2 & 63.4 & 6.6
& 66.7 & 61.7 & 15.0
& \textbf{83.9} & \textbf{83.9} & \textbf{0.0}
& 51.7 & 50.0 & \textbf{8.3}
& 58.8 & 58.8 & \textbf{2.0} \\

$\pi_{0.5}$ (MOSAIC)
& \textbf{74.9} & \textbf{72.7} & \textbf{6.2}
& \textbf{73.3} & \textbf{70.0} & \textbf{11.7}
& \textbf{83.9} & \textbf{83.9} & \textbf{0.0}
& \textbf{76.7} & \textbf{71.7} & 10.0
& \textbf{64.7} & \textbf{64.7} & \textbf{2.0} \\

\bottomrule
\end{tabular} \vspace{-6mm}
\end{table*}

\subsection{Experiments on Real-Robot}

\begin{figure}[t] \vspace{-1mm}
    \centering
    \includegraphics[width=0.85\linewidth]{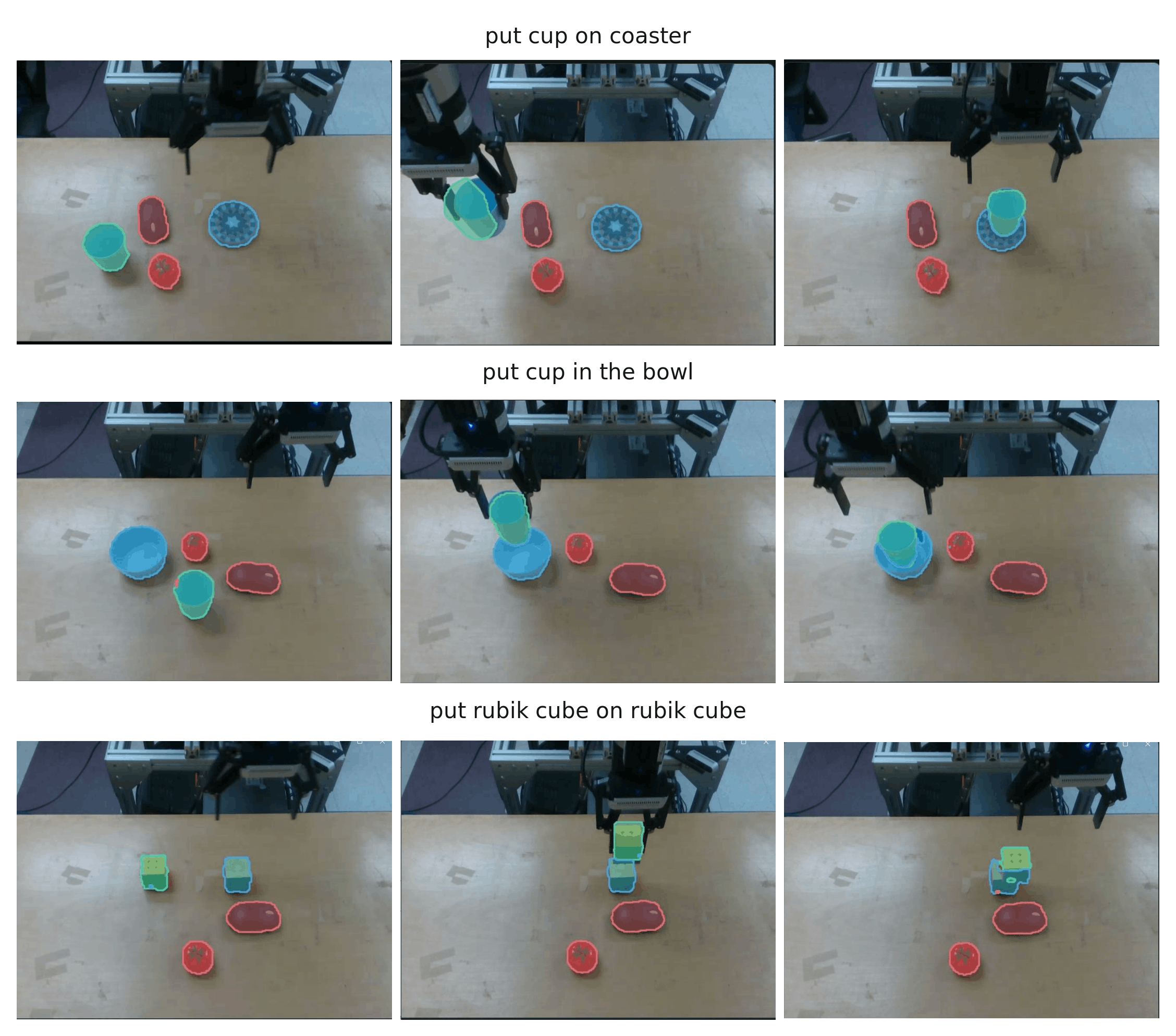}\vspace{-2mm}
    \caption{Real-world experiments. The model is able to perform different tasks while generating the masks and conditioning on them during action generation.}
    \label{fig:real-robot} \vspace{-6mm}
\end{figure}

For real-world experiments, we first fully fine-tune the base $\pi_{0.5}$ policy on our real-world dataset without auxiliary tasks, and then fine-tune MOSAIC on the same data. We test the resulting policy across different manipulation tasks, with qualitative results shown in Fig.~\ref{fig:real-robot} and the supplementary videos. 
Across these experiments, our model identifies the target object, destination region, and surrounding obstacles and use this information throughout action generation. This enables the policy to maintain task-relevant spatial awareness while executing manipulation trajectories in the presence of nearby distractors and potential collision regions.
The real-world results therefore show that the proposed modules transfer beyond simulation and can improve the spatial grounding of a pretrained VLA for manipulation in cluttered physical environments.

\section{CONCLUSIONS}

We proposed MOSAIC, a modular spatial- and obstacle-aware framework that augments pretrained VLA policies with role-specific semantic grounding, object-level geometric tokens, and obstacle-aware attention guidance. These complementary cues incorporate task-relevant geometry and potential robot-obstacle interactions into action generation. Evaluations on RoboPRO and experiments on the real-world tasks demonstrate improved robustness and performance in cluttered environments. Future work will explore distilling object-level 3D shape representations into the VLA backbone to strengthen its intrinsic geometric understanding and support spatial reasoning and generalization.

\textbf{Acknowledgment} we used AI-generated images in Fig.~\ref{fig:general} and \ref{fig:main}. GPT-6 generated the robot and objects images.




\bibliographystyle{IEEEtran}
\bibliography{references}

\end{document}